\documentclass[11pt,a4paper]{article}
\usepackage[hyperref]{emnlp2019}
\usepackage{times}
\usepackage{graphicx}
\usepackage{tikz}
\usepackage{latexsym}
\usepackage{pgfplots}
\usepackage{booktabs}
\usepackage{amsmath}
\usepackage{url}

\aclfinalcopy

\newcommand{\ignore}[1]{}
\newcommand*\samethanks[1][\value{footnote}]{\footnotemark[#1]}

\title{Data Augmentation for BERT Fine-Tuning\\ in Open-Domain Question Answering}

\author{Wei Yang,$^{1,2}$\thanks{\hspace{0.25cm}equal contribution}\hspace{0.2cm} Yuqing Xie,$^{1,2}$\samethanks\hspace{0.2cm} {\bf Luchen Tan},$^2$
{\bf Kun Xiong,}$^2$ {\bf Ming Li,}$^{1,2}$ \and {\bf Jimmy Lin}$^{1,2}$\vspace{0.1cm}\\
$^1$ David R. Cheriton School of Computer Science, University of Waterloo \\
$^2$ RSVP.ai}

\date{}

\begin{document}

\maketitle

\begin{abstract}
Recently, a simple combination of passage retrieval using off-the-shelf IR techniques and a BERT reader was found to be very effective for question answering directly on Wikipedia, yielding a large improvement over the previous state of the art on a standard benchmark dataset.
In this paper, we present a data augmentation technique using distant supervision that exploits positive as well as negative examples.
We apply a stage-wise approach to fine tuning BERT on multiple datasets, starting with data that is ``furthest'' from the test data and ending with the ``closest''.
Experimental results show large gains in effectiveness over previous approaches on English QA datasets, and we establish new baselines on two recent Chinese QA datasets.
\end{abstract}

\section{Introduction}

BERT \cite{devlin2018bert} represents the latest refinement in a series of neural models that take advantage of pretraining on a language modeling task~\cite{N18-1202,radford2018improving}.
Researchers have demonstrated impressive gains in a broad range of NLP tasks, from sentence classification to sequence labeling.
Recently, \citet{yang2019end} showed that combining a BERT-based reader with passage retrieval using the Anserini IR toolkit yields a large improvement in question answering directly from a Wikipedia corpus, measured in terms of exact match on a standard benchmark~\cite{P17-1171}.

Interestingly, the approach of \citet{yang2019end} represents a simple method to combining BERT with off-the-shelf IR.
In this paper, we build on these initial successes to explore how much further we can push this simple architecture by data augmentation, taking advantage of distant supervision techniques to gather more and higher-quality training data to fine tune BERT.
Experiments show that, using the same reader model as \citet{yang2019end}, our simple data-augmentation techniques yield additional large improvements.
To illustrate the robustness of our methods, we also demonstrate consistent gains on another English QA dataset and present baselines for two additional Chinese QA datasets (which have not to date been evaluated in an ``end-to-end'' manner).

In addition to achieving state-of-the-art results, we contribute important lessons on how to leverage BERT effectively for question answering.
First, most previous work on distant supervision focuses on generating positive examples, but we show that using existing datasets to identify negative training examples is beneficial as well.
Second, we propose an approach to fine-tuning BERT with disparate datasets that works well in practice:\ our heuristic is to proceed in a stage-wise manner, beginning with the dataset that is ``furthest'' from the test data and ending with the ``closest''.

\vspace{-0.05cm}
\section{Background and Related Work}
\vspace{-0.05cm}

In this paper, we tackle the ``end-to-end'' variant of the question answering problem, where the system is only provided a large corpus of articles.
This stands in contrast to reading comprehension datasets such as SQuAD~\cite{D16-1264}, where the system works with a single pre-determined document, or most QA benchmarks today such as TrecQA~\cite{Yao13answerextraction}, Wiki\-QA~\cite{yang2015wikiqa}, and MSMARCO~\cite{nguyen2016ms}, where the system is provided a list of candidate passages to choose from.
This task definition, which combines a strong element of information retrieval, traces back to the Text Retrieval Conferences (TRECs) in the late 1990s~\cite{Voorhees_Tice_TREC8}, but there is a recent resurgence of interest in this formulation~\cite{P17-1171}.

The roots of the distant supervision techniques we use trace back to at least the 1990s~\cite{P95-1026,Riloff:1996:AGE:1864519.1864542}, although the term had not yet been coined.
Such techniques have recently become commonplace, especially as a way to gather large amounts of labeled examples for data-hungry neural networks and other machine learning algorithms.
Specific recent applications in question answering include \citet{bordes2015large}, \citet{P17-1171}, \citet{P18-1161}, as well as \citet{P17-1147} for building benchmark test collections.

\section{Approach}

In this work, we fix the underlying model and focus on data augmentation techniques to explore how to best fine-tune BERT.
We use the same exact setup as the ``paragraph'' variant of BERTserini~\cite{yang2019end}, where the input corpus is pre-segmented into paragraphs at index time, each of which is treated as a ``document'' for retrieval purposes.
The question is used as a ``bag of words'' query to retrieve the top $k$ candidate paragraphs using BM25 ranking.
Each paragraph is then fed into the BERT reader along with the original natural language question for inference.
Our reader is built using Google's reference implementation, but with a small tweak:\ to allow comparison and aggregation of results from different segments, we remove the final softmax layer over different answer spans; cf.~\cite{P18-1078}.
For each candidate paragraph, we apply inference over the entire paragraph, and the reader selects the best text span and provides a score.
We then combine the reader score with the retriever score via linear interpolation: $S = (1- \mu) \cdot S_{\textrm{Anserini}} + \mu \cdot S_{\textrm{BERT}}$, 
where $\mu \in [0,1]$ is a hyperparameter (tuned on a training sample).

One major shortcoming with BERTserini is that \citet{yang2019end} only fine tune on SQuAD, which means that the BERT reader is exposed to an impoverished set of examples; all SQuAD data come from a total of only 442 documents.
This contrasts with the diversity of paragraphs that the model will likely encounter at inference time, since they are selected from potentially millions of articles.
The solution to this problem, of course, is to fine tune BERT with the types of paragraphs it is likely to see at inference time.
Unfortunately, such data does not exist for modern QA test collections.
Distant supervision can provide a bridge.

Starting from a source dataset comprising question--answer pairs (for example, SQuAD), we can create training data for a specific corpus by using passage retrieval to fetch paragraphs from that corpus (with the question as the query) and then searching (i.e., matching) for answer instances in those paragraphs.
A hyperparameter here is $n$, the number of candidates we examine from passage retrieval.
Larger values of $n$ will lead to more training examples, but as $n$ increases, so does the chance that a paragraph will spuriously match the answer without actually answering the question.

The above technique allows us to extract positive training examples, but previous work has shown the value of negative examples, specifically for QA~\cite{Zhang:2017:AEH:3077136.3080645}.
To extract negative examples, we sample the top $n$ candidates from passage retrieval for paragraphs that do not contain the answer, with a ratio of $d$:1.
That is, for every positive example we find, we sample $d$ negative examples, where $d$ is also a hyperparameter.
Note that these negative examples are also noisy, since they may in fact contain an alternate correct (or acceptable) answer to the question, one that differs from the answer given in the source dataset.

Thus, given a corpus, we can create using distant supervision a new dataset that is specifically adapted to a particular passage retrieval method.
For convenience, we refer to training data gathered using this technique that only contain positive examples as DS($+$) and use DS($\pm$) to refer to the additional inclusion of negative examples.

Next, we have a design decision regarding how to fine tune BERT using the source QA pairs (SRC) and the augmented dataset using distant supervision (DS).
There are three possibilities:

\smallskip \noindent \textbf{SRC + DS}:\
Fine tune BERT with all data, grouped together.
In practice, this means that the source and augmented data are shuffled together.

\smallskip \noindent \textbf{DS $\rightarrow$ SRC}:\
Fine tune the reader on the augmented data and then the source dataset.

\smallskip \noindent \textbf{SRC $\rightarrow$ DS}:\
Fine tune the reader on the source dataset and then the augmented data.

\smallskip \noindent
Experiment results show that of the three choices above, the third option is the most effective.
More generally, when faced with multiple, qualitatively-different datasets, we advocate a stage-wise fine-tuning strategy that starts with the dataset ``furthest'' to the task at hand and ending with the dataset ``closest''.

Another way to think about using different datasets is in terms of a very simple form of transfer learning.
The stage-wise fine-tuning strategy is in essence trying to transfer knowledge from labeled data that is not drawn from the same distribution as the test instances.
We wish to take advantage of transfer effects, but limit the scope of erroneous parameterizations.
Thus it makes sense not to intermingle qualitatively different datasets, but to fine tune the model in distinct stages.

\section{Experimental Setup}
\label{sec:setup}

To show the generalizability of our data augmentation technique, we conduct experiments on two English datasets:\ SQuAD (v1.1) and Trivia\-QA~\cite{P17-1147}.
For both, we use the 2016-12-21 dump of English Wikipedia, following \citet{P17-1171}.
We also examine two Chinese datasets:\ CMRC \cite{cui2018span} and DRCD~\cite{shao2018drcd}.
For these, we use the 2018-12-01 dump of Chinese Wikipedia, tokenized with Lucene's \texttt{CJKAnalyzer} into overlapping bigrams.
We apply hanziconv\footnote{https://pypi.org/project/hanziconv/0.2.1/} to transform the corpus into simplified characters for CMRC and traditional characters for DRCD.

Following \citet{yang2019end}, to evaluate answers in an end-to-end setup, we disregard the paragraph context from the original datasets and use only the answer spans.
As in previous work, exact match (EM) score and F$_1$ score (at the token level) serve as the two primary evaluation metrics.
In addition, we compute recall (R), the fraction of questions for which the correct answer appears in {\it any} retrieved paragraph; to make our results comparable to \citet{yang2019end}, Anserini returns the top $k=100$ paragraphs to feed into the BERT reader.
Note that this recall is {\it not} the same as the token-level recall component in the F$_1$ score.
Statistics for the datasets are shown in Table~\ref{tab:stats}.\footnote{Note the possibly confusing terminology:\ for SQuAD (as well as the other datasets), what we use for test is actually the publicly-available development set (same as previous work).}

\begin{table}[t]
\centering\centering\resizebox{\columnwidth}{!}{
\begin{tabular}{lrrrr}
\toprule
			& SQuAD 	& TriviaQA			&  CMRC		& DRCD \\ \toprule
Train			& 87,599			& 87,622					&10,321		& 26,936\\ 
Test			& 10,570			& 11,313					& 3,351		& 3,524\\ 
\midrule
DS($+$)		& 118,406		& 264,192	 	& 10,223	& 41,792 \\ 
DS($\pm$)	& 710,338		& 789,089		& 71,536 	& 246,604\\ \bottomrule
\end{tabular}}
\label{tab:stats}
\caption{Number of examples in each dataset. A example means a paragraph-question pair.}
\end{table}

For data augmentation, based on preliminary experiments, we find that examining $n=10$ candidates from passage retrieval works well, and we further discover that effectiveness is insensitive to the amount of negative samples.
Thus, we eliminate the need to tune $d$ by simply using all passages that do not contain the answer as negative examples.
The second block of Table~\ref{tab:stats} shows the sizes of the augmented datasets constructed using our distant supervision techniques:\ DS($+$) contains positive examples only, while DS($\pm$) includes both positive and negative examples.

There are two additional characteristics to note about our data augmentation techniques:\
The most salient characteristic is that SQuAD, CMRC, and DRCD all have source answers drawn from Wikipedia (English or Chinese), while TriviaQA includes web pages as well as Wikipedia.
Therefore, for the first three collections, the source and augmented datasets share the same document genre---the primary difference is that data augmentation increases the amount and diversity of answer passages seen by the model during training.
For TriviaQA, however, we consider the source and augmented datasets as coming from different genres (noisy web text vs.\ higher quality Wikipedia articles).
Furthermore, the TriviaQA augmented dataset is also much larger---suggesting that those questions are qualitatively different (e.g., in the manner they were gathered).
These differences appear to have a substantial impact, as experiment results show that TriviaQA behaves differently than the other three collections.

For model training, we begin with the BERT-Base model (uncased, 12-layer, 768-hidden, 12-heads, 110M parameters), which is then fine-tuned using the various conditions described in the previous section.
All inputs to the model are padded to 384 tokens; the learning rate is set to $3 \times 10^{-5}$ and all other defaults settings are used.

\section{Results}

\begin{table}[t]
\centering\resizebox{\columnwidth}{!}{
\begin{tabular}{lccc}
\toprule
Model 					& EM 	& F$_1$		& R  \\ 
\toprule
Dr.QA \cite{P17-1171} 		& 27.1 	& - 		& 77.8 	\\
Dr.QA + Fine-tune 			& 28.4 	& -  		& - \\
Dr.QA + Multitask 			& 29.8 	& - 		&-  \\
$\text{R}^3$ \cite{wang2017r} 	& 29.1 	& 37.5	& -  \\
\citet{D18-1055}  			& 29.8  	& - 		&-  \\
Par. R. \cite{D18-1053}  		& 28.5 	& - 		& 83.1 \\
Par. R. + Answer Agg. 		& 28.9 	& - 		&-  \\
Par. R. + Full Agg. 			& 30.2 	& - 		&-  \\
\textsc{Minimal}~\cite{P18-1160}	& 34.7 	& 42.5 	& 64.0	 \\
\midrule
BERTserini \cite{yang2019end} & 38.6	& 46.1	& 85.9	\\
\midrule
SRC						& 41.8	& 49.5	& 85.9	 \\
DS($+$)					& 44.0	& 51.4 	& 85.9	\\
DS($\pm$)				& 48.7	& 56.5	& 85.9	\\
\midrule
SRC + DS($\pm$)			& 45.7	& 53.5	& 85.9 	 \\
DS($\pm$) $\rightarrow$ SRC	& 47.4	& 55.0	& 85.9	 \\
SRC $\rightarrow$ DS($\pm$)	& \textbf{50.2}& \textbf{58.2}& 85.9 \\
\bottomrule
\end{tabular}}
\caption{Results on \text{SQuAD}}
\label{table:main}
\end{table}

Our main results on SQuAD are shown in Table~\ref{table:main}.
The row marked ``SRC'' indicates fine tuning with SQuAD data only and matches the BERTserini condition of~\citet{yang2019end}; we report higher scores due to engineering improvements (primarily a Lucene version upgrade).
As expected, fine tuning with augmented data improves effectiveness, and experiments show that while training with positive examples using DS($+$) definitely helps, an even larger boost comes from leveraging negative examples using DS($\pm$).
In both these cases, we only fine tune BERT with the augmented data, ignoring the source data.

What if we use the source data as well?
Results show that ``lumping'' all training data together (both the source and augmented data) to fine tune BERT is not the right approach:\ in fact, the SRC + DS($\pm$) condition performs worse than just using the augmented data alone.
Instead, disparate datasets should be leveraged using the stage-wise fine-tuning approach we propose, according to our heuristic of starting with the dataset that is ``furtherest'' away from the test data.
That is, we wish to take advantage of all available data, but the last dataset we use to fine tune BERT should be ``most like'' the test data the model will see at inference time.
Indeed, this heuristic is borne out empirically, as SRC $\rightarrow$ DS($\pm$) yields another boost over using DS($\pm$) only.
Further confirmation for this heuristic comes from an alternative where we switch the order of the stages, DS($\pm$) $\rightarrow$ SRC, which yields results worse than DS($\pm$) alone.
We note that our best configuration beats BERTserini, the previous state of the art, by over ten points.
Note that recall in all our conditions is the same since we are not varying the passage retrieval algorithm, and in each case Anserini provides exactly the same candidate passages.
Improvements come solely from a better BERT reader.

\begin{table}[t]
\centering\resizebox{\columnwidth}{!}{
\begin{tabular}{lccc}
\toprule
Model 					& EM 	& F$_1$  	&R	 	\\
\midrule
$\text{R}^3$ \cite{wang2017r} 	& 47.3 	&53.7	&-	\\
DS-QA~\cite{P18-1161}			& 48.7	& 56.3	&-	\\
Evidence Agg.~\cite{Shuohang2017} &50.6 &57.3&-\\
\midrule
SRC 					&51.0	& 56.3	&83.7	\\
DS($+$) 					&48.2	& 53.6	&83.7	\\
DS($\pm$) 				&\textbf{54.4}	& \textbf{60.2}	&83.7\\
\midrule
SRC + DS($\pm$)			& 53.1 & 58.6 & 83.7	 \\
DS($\pm$) $\rightarrow$ SRC  & 49.8 	& 55.9	& 83.7 \\
SRC $\rightarrow$ DS($\pm$) 	&53.7	&59.3	&83.7\\
\bottomrule
\end{tabular}}
\caption{Results on TriviaQA}
\label{table:trivia}
\end{table}

Results on TriviaQA are shown in Table~\ref{table:trivia}.
With just fine tuning on the source dataset, we obtain a score that is only slightly above the previous state of the art~\cite{Shuohang2017}.
Interestingly, using only positive examples leads to worse effectiveness than just using the source dataset.
However, fine tuning on both positive and negative examples leads to a three point boost in exact match score, establishing a new high score on this dataset.

Experiments on fine tuning with both source and augmented data show the same pattern as with SQuAD:\ stage-wise tuning is more effective than just combining datasets, and tuning should proceed in the ``furthest to closest'' sequence we propose.
While data augmentation no doubt helps (beats the source-only baseline), for this dataset the highest effectiveness is achieved by disregarding the source dataset completely; that is, DS($\pm$) beats SRC $\rightarrow$ DS($\pm$).
We attribute this behavior to the difference between TriviaQA and the other datasets discussed in Section~\ref{sec:setup}:\ it appears that gains from transfer effects are outweighed by genre mismatch.

Results on the Chinese datasets are shown in Table~\ref{table:chinese}.
To our knowledge, they have only been evaluated as reading comprehension tests, not in the ``end-to-end'' setup that we tackle here (requiring retrieval from a sizeable corpus).
Although there is no previous work to compare against, our results provide a strong baseline for future work.

\begin{table}[t]
\centering\begin{tabular}{lccc}
\toprule
Model 					& EM 	& F$_1$ 	& R 	\\
\toprule
CMRC \\
~SRC            				& 44.5	&60.9       & 86.5 \\ 
~DS($+$)       				& 45.5	&61.1  	& 86.5\\
~DS($\pm$)     				& 48.3	&63.9  	& 86.5\\
~SRC + DS($\pm$)			& 49.0	& 64.6	& 86.5	 \\
~DS($\pm$) $\rightarrow$ SRC  &45.6 & 61.9 & 86.5 \\
~SRC $\rightarrow$ DS($\pm$)	& \textbf{49.2}&\textbf{65.4}& 86.5 \\
\midrule
DRCD\\
~SRC      					& 50.7	&65.0       & 81.5   \\ 
~DS($+$) 					&50.5	&64.3       & 81.5 \\ 
~DS($\pm$)				& 53.2	&66.0       & 81.5\\ 
~SRC + DS ($\pm$)			& \textbf{55.4}	& \textbf{67.7}	& 81.5	 \\
~DS($\pm$) $\rightarrow$ SRC  &53.4	& 67.1	& 81.5 \\
~SRC $\rightarrow$ DS($\pm$) &54.4 	&67.0	& 81.5\\
\bottomrule
\end{tabular}
\caption{Results on the two Chinese datasets:\ CMRC (top) and DRCD (bottom).}
\label{table:chinese}
\end{table}

Experiment results on the two Chinese datasets support the same conclusions as SQuAD:\
First, we see that data augmentation using distant supervision is effective.
Second, including both positive and negative training examples is better than having positive examples only.
Third, when leveraging multiple datasets, our ``furthest to closest'' heuristic for stage-wise tuning yields the best results.
Since the source datasets also draw from (Chinese) Wikipedia, we benefit from fine tuning with both source and augmented data.

\section{Conclusions}

In this paper, we have further advanced the state of the art in end-to-end open-domain question answering using simple BERT models.
We focus on data augmentation using distant supervision techniques to construct datasets that are closer to the types of paragraphs that the reader will see at inference time.
Explained this way, it should not come as a surprise that effectiveness improves as a result.
This work confirms perhaps something that machine learning practitioners already know too well:\ quite often, the best way to better results is not better modeling, but better data preparation.


\end{document}